\documentclass[sigconf,nonacm]{acmart}
\AtBeginDocument{%
  }

\setcopyright{acmlicensed}
\copyrightyear{}
\acmYear{}
\acmDOI{}

\usepackage{pifont}
\newcommand{\cmark}{\ding{51}}%
\newcommand{\xmark}{\ding{55}}%
\usepackage{threeparttable}
\usepackage{multirow}
\usepackage{xcolor}
\usepackage{outline}
\usepackage{enumitem}

\begin{document}

\title{Multi-modal Time Series Analysis: A Tutorial and Survey}

\author{
Yushan Jiang\textsuperscript{\rm 1*},
Kanghui Ning\textsuperscript{\rm 1*},
Zijie Pan\textsuperscript{\rm 1*},
Xuyang Shen\textsuperscript{\rm 1},
Jingchao Ni\textsuperscript{\rm 2},
Wenchao Yu\textsuperscript{\rm 4},
Anderson Schneider\textsuperscript{\rm 3},
Haifeng Chen\textsuperscript{\rm 4$\dag$},
Yuriy Nevmyvaka\textsuperscript{\rm 3$\dag$},
Dongjin Song\textsuperscript{\rm 1$\dag$}}
\affiliation{%
  \institution{\textsuperscript{\rm 1}University of Connecticut \hspace{0.4em}
  \textsuperscript{\rm 2}University of Houston
  }
  \textsuperscript{\rm 3}Morgan Stanley \hspace{0.4em}
  \textsuperscript{\rm 4}NEC Laboratories America 
  \city{} 
  \state{}
  \country{}
  }

\renewcommand{\shortauthors}{Jiang, et al.}
\newcommand\blfootnote[1]{%
  \begingroup
  \renewcommand\thefootnote{}\footnote{#1}%
  \addtocounter{footnote}{-1}%
  \endgroup
}
\begin{abstract}
Multi-modal time series analysis has recently emerged as a prominent research area in data mining, driven by the increasing availability of diverse data modalities, such as text, images, and structured tabular data from real-world sources. However, effective analysis of multi-modal time series is hindered by data heterogeneity, modality gap, misalignment, and inherent noise. Recent advancements in multi-modal time series methods have exploited the multi-modal context via cross-modal interactions based on deep learning methods, significantly enhancing various downstream tasks. In this tutorial and survey, we present a systematic and up-to-date overview of multi-modal time series datasets and methods. We first state the existing challenges of multi-modal time series analysis and our motivations, with a brief introduction of preliminaries. Then, we summarize the general pipeline and categorize existing methods through a unified cross-modal interaction framework encompassing fusion, alignment, and transference at different levels (\textit{i.e.}, input, intermediate, output), where key concepts and ideas are highlighted. We also discuss the real-world applications of multi-modal analysis for both standard and spatial time series, tailored to general and specific domains. Finally, we discuss future research directions to help practitioners explore and exploit multi-modal time series. The up-to-date resources are provided in the GitHub repository\blfootnote{* equal contribution. $^{\dag}$ Yuriy Nevmyvaka, Haifeng Chen and Dongjin Song are the corresponding authors.}\footnote{\href{https://github.com/UConn-DSIS/Multi-modal-Time-Series-Analysis}{https://github.com/UConn-DSIS/Multi-modal-Time-Series-Analysis}}.

\end{abstract}






\maketitle

\section{Introduction}
\begin{figure}
    \centering
    \includegraphics[width=0.98\linewidth]{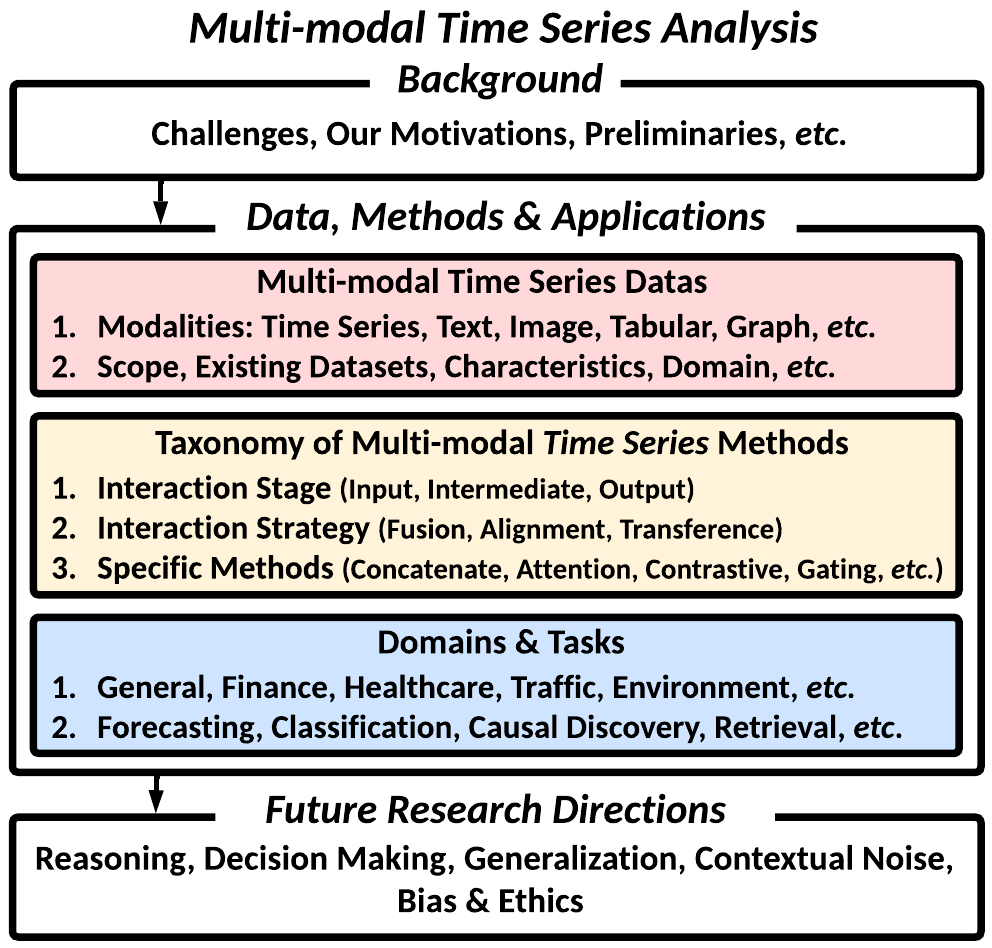}
    \caption{The framework of our tutorial and survey.}
    \label{fig:framework}
    \vspace{-3mm}
\end{figure}
Time series analysis is a fundamental task in data mining, driven by the proliferation of sequential data exhibiting rich temporal dynamics across diverse real-world systems. With the advent of deep learning, various methods have been proposed to effectively model complex temporal relationships within time series~\cite{qin2017dual,mtgnn,wutimesnet,nietime,chentsmixer,pan2024s,yi2024frequency}, facilitating downstream tasks in diverse domains, including healthcare~\cite{jin2018treatment,zhang2020adversarial,feuerriegel2024causal}, finance~\cite{zhang2017stock,rezaei2021stock}, transportation~\cite{guo2019attention,zhang2021traffic,ji2023spatio} and environmental sciences~\cite{chen2022physics,chen2023physics}.

In practice, time series are often associated with external contexts beyond their temporal dynamics~\cite{williams2024context,chattopadhyay2025contextmattersleveragingcontextual}. Such contexts are multi-modal, encompassing a variety of representations, such as texts~\cite{lee2024moat,2024wangnewsforecast}, images~\cite{ekambaram2020attention,skenderi2024well}, tables~\cite{chattopadhyay2025contextmattersleveragingcontextual}, and graphs~\cite{sawhney-etal-2020-deep}, which carry rich semantic information for time series analysis. As such, incorporating the multi-modal contexts allows models to have a comprehensive view of underlying systems, capture subtle dependencies, and explain complex temporal behaviors more accurately.  

Effective analysis of multi-modal time series, however, is hindered by several key challenges in terms of {\em data heterogeneity}, {\em modality gap} and {\em contextual relevance}. 
First, different modalities exhibit distinct statistical properties, structures, and dimensionalities, leading to discrepancies in feature distributions and semantic meanings. For instance, while time series data is sequentially ordered with temporal dependencies, textual and image data contains rich contextual semantics and correlations. Aligning these heterogeneous data into a unified representation space is 
non-trivial. Second, the textual, tabular, or visual contexts may appear at different timesteps or granularities. Such temporal misalignment may impede meaningful cross-modal interactions. Third, real-world data is inevitably noisy with irrelevant information that may mislead correlation learning, resulting in suboptimal performance. For example, in finance, news articles related to stock market prediction often contain much redundant or speculative narratives that does not reflect actual market conditions. Therefore, the focus of multi-modal time series analysis is to effectively capture complementary and relevant information from multi-modal context and leverage it for predictive or analytical tasks.

More recently, an increasing number of multi-modal methods have shown promise in exploiting contextual information from diverse data sources, which boosts performance in wide tasks ranging from forecasting~\cite{lee2024moat,liu2024timecma}, classification~\cite{lee2024timecap,li2023frozen}, anomaly detection~\cite{XING2023108567} to retrieval~\cite{latentspaceprojection} and causal discovery~\cite{shen2024exploringmultimodalintegrationtoolaugmented,zheng2024mulan}.
Despite the promising results of multi-modal time series methods, they are tailored for their own tasks with domain-specific applications.
The existing literature lacks a comprehensive and systematic review that provides a unified perspective on the underlying principles and pipelines for multi-modal time series learning.
In this survey, we provide a systematic and up-to-date overview of existing methods for multi-modal time series analysis. As shown in Figure~\ref{fig:framework}, 
we discuss the challenges, motivations, and preliminaries of multi-modal time series. Then we introduce the general pipeline for multi-modal time series analysis and propose three types of interactions for cross-modal modeling between time series and other modalities --- {\em fusion}, {\em alignment}, and {\em transference} --- at the input, intermediate and output level, respectively. We also discuss the applications of multi-modal time series across multiple domains. Furthermore, we provide Table~\ref{tab:taxonomy} to comprehensively summarize representative methods, encapsulating the modalities, fine-grained cross-modal interactions, real-world domains and tasks. Finally, we highlight potential future research opportunities to further advance time series analysis with multi-modal data.
In summary, the major contributions of our survey are:
\begin{itemize}[leftmargin=*]
\item We systematically catalog over 40 multi-modal time series methods with the corresponding open-source datasets.
\item We uniquely categorize the existing methods into a unified cross-modal interaction framework, highlighting fusion, alignment, and transference at the input/intermediate/output levels.
\item We discuss real-world applications of multi-modal time series and identify promising future directions, encouraging researchers and practitioners to explore and exploit multi-modal time series.
\end{itemize}

\section{Background and Our Scope}
\subsection{Multi-modal Machine Learning}
Recent advancements in multi-modal machine learning have significantly enhanced models' ability to process and integrate data from diverse modalities, such as language, acoustic, vision, and tabular data~\cite{hager2023best,ruan2023mm,zhang2024vision}. With the development of deep learning architectures and sophisticated interaction designs, models are able to learn, infer, and reason by integrating multiple communicative modalities. Current research in multi-modal machine learning spans multiple key areas, including (1) representing multi-modal data to encode joint and individual characteristics, (2) identifying interconnections between modality elements, (3) transferring knowledge across modalities, and (4) theoretically and empirically analyzing the underlying learning process in a quantitative manner. We refer the audiences to the recent surveys~\cite{baltruvsaitis2018multimodal,liang2024foundations} for a more detailed overview of general multi-modal machine learning research. Building upon these advancements, we investigate multi-modal time series analysis with a focus on modeling temporal dependencies and leveraging the data interactions across heterogeneous modalities for predictive and analytical tasks.

\subsection{Multi-modal Time Series Analysis}

Multi-modal time series analysis aims to model time series data in combination with other complementary modalities. By leveraging cross-modal interactions, this approach yields deeper insights and more robust solutions for a wide range of predictive and analytical tasks across diverse real-world contexts.

This survey aims to provide a unique and systematic perspective on \textit{effectively leveraging cross-modal interactions from relevant real-world contexts to advance multi-modal time series analysis}, addressing both foundational principles and practical solutions. Our assessment is threefold: (1) reviewing multi-modal time series data (Section \ref{sec:multimodal_time_series_data}), (2) analyzing cross-modal interactions between time series and other modalities (Section 4), and (3) revealing the impact of multi-modal time series analysis in applications across diverse domains (Section 5). 

To resolve ambiguities, we define the scope of our survey by clarifying the types of time series considered and the criteria for multi-modal time series methods. First, we mainly consider standard time series and spatial time series. For the latter, spatial structures (often represented as graphs) are inherently paired with temporal data rather than treated as a separate modality. Second, we focus on methods that leverage \textbf{multi-modal inputs} from real-world contexts to provide complementary information, but for generation and retrieval tasks, the focus is more on transforming the input modality to another output modality. We acknowledge recent research on representing time series as \textbf{a single modality} (\textit{e.g.,} time series as images~\cite{li2023time,prithyani2024feasibility,daswani2024plots,zhou2024can,zhuang2024see}, time series as tabular data~\cite{hoo2025tabular}) for downstream tasks. However, as these approaches are less relevant to our scope, we refer readers to their respective works.

Besides, we would like to highlight the difference between our survey and recent related survey and position papers. 
Ni \textit{et al.}~\cite{ni2025harnessing} focuses on imaging-based transformations of time series and subsequent visual modeling techniques, where the discussion on multi-modal models is limited to those involving vision modalities. Kong \textit{et al.}~\cite{kong2025position} concentrates on the use of multi-modal large language models (LLMs) for enhancing reasoning capabilities (\textit{e.g.,} causal reasoning, QA, planning, \textit{etc.}) with multi-modal context. In contrast, our survey provides a broader and structured framework by delivering a systematic and unified perspective of multi-modal time series analysis, not limited to a specific modality or task type.

\section{Multi-modal Time Series Data}\label{sec:multimodal_time_series_data}

\begin{table*}[t]
  \setlength{\tabcolsep}{4.3pt}
  \renewcommand{\arraystretch}{0.83}
  \footnotesize
\centering
\caption{Representative open-source multi-modal time series datasets and across domains.}
\label{tab:mm_datasets}
\begin{tabular}{lll}
\toprule
\textbf{Domain} & \textbf{Dataset} {(Superscripts include the URLs to the datasets)} & \textbf{Modalities} \\
\midrule
Healthcare & MIMIC-III~\cite{Johnson2016}\textsuperscript{\href{https://mimic.mit.edu/docs/iii/}{[1]}}, MIMIC-IV~\cite{Johnson2021}\textsuperscript{\href{https://physionet.org/content/mimiciv/1.0/}{[2]}} & TS, Text, Tabular \\
& ICBHI~\cite{Rocha2019}\textsuperscript{\href{https://bhichallenge.med.auth.gr/ICBHI_2017_Challenge}{[3]}}, Coswara~\cite{Bhattacharya2023}\textsuperscript{\href{https://zenodo.org/records/7188627}{[4]}}, KAUH~\cite{KAUHdata}\textsuperscript{\href{https://data.mendeley.com/datasets/jwyy9np4gv/3}{[5]}}, PTB-XL~\cite{Wagner2020}\textsuperscript{\href{https://physionet.org/content/ptb-xl/1.0.3/}{[6]}}, ZuCo~\cite{ZuCo2019, ZuCo2020}\textsuperscript{\href{https://github.com/norahollenstein/zuco-benchmark}{[7]}}  & TS, Text \\
& Image-EEG~\cite{ImageEEG}\textsuperscript{\href{https://github.com/gifale95/eeg_encoding}{[8]}} & TS, Image \\ 

\midrule
Finance & FNSPID~\cite{FNSPID}\textsuperscript{\href{https://github.com/Zdong104/FNSPID_Financial_News_Dataset}{[9]}}, ACL18~\cite{ACL18}\textsuperscript{\href{https://github.com/yumoxu/stocknet-dataset}{[10]}}, CIKM18~\cite{CIKM18}\textsuperscript{\href{https://github.com/wuhuizhe/CHRNN}{[11]}}, DOW30~\cite{chen2023chatgpt}\textsuperscript{\href{https://github.com/ZihanChen1995/ChatGPT-GNN-StockPredict}{[12]}} & TS, Text \\
\midrule
Multi-domain & Time-MMD~\cite{liu2024timemmd}\textsuperscript{\href{https://github.com/AdityaLab/Time-MMD}{[13]}}, TimeCAP~\cite{lee2024timecap}\textsuperscript{\href{https://github.com/ameliawong1996/From_News_to_Forecast}{[14]}}, NewsForecast~\cite{2024wangnewsforecast}\textsuperscript{\href{https://github.com/geon0325/TimeCAP}{[15]}}, TTC~\cite{kim2024multimodalforecasterjointlypredicting}\textsuperscript{\href{https://github.com/Rose-STL-Lab/Multimodal_Forecasting}{[16]}}, CiK~\cite{williams2024context}\textsuperscript{\href{https://servicenow.github.io/context-is-key-forecasting/v0/}{[17]}}, TSQA~\cite{kong2025timemqatimeseriesmultitask}\textsuperscript{\href{https://huggingface.co/Time-QA}{[18]}} & TS, Text\\
\midrule
Retail & VISUELLE~\cite{skenderi2024well}\textsuperscript{\href{https://github.com/HumaticsLAB/GTM-Transformer}{[19]}} & TS, Image, Text  \\
\midrule
IoT & LEMMA-RCA~\cite{zheng2024lemmarca}\textsuperscript{\href{https://lemma-rca.github.io/}{[20]}} & TS, Text \\
\midrule
Speech & LRS3~\cite{2018LRS3}\textsuperscript{\href{https://www.robots.ox.ac.uk/~vgg/data/lip_reading/}{[21]}}, VoxCeleb2~\cite{Chung_2018}\textsuperscript{\href{https://www.robots.ox.ac.uk/~vgg/data/voxceleb/}{[22]}} & TS (Audio), Image  \\
\midrule
Traffic & NYC-taxi, NYC-bike~\cite{Li2024UrbanGPTSL}\textsuperscript{\href{https://huggingface.co/datasets/bjdwh/ST_data_urbangpt}{[23]}} & ST, Text \\
\midrule
Environment & Terra~\cite{NEURIPS2024_Terra}\textsuperscript{\href{https://github.com/CityMind-Lab/NeurIPS24-Terra}{[24]}} & ST, Text \\
\bottomrule
\end{tabular}
\end{table*}

\subsection{Modalities in Multi-modal Time Series Data}
Multi-modal time series data often originate from diverse sources, each exhibiting unique characteristics that influence how they are processed and analyzed. Besides \textbf{Time Series}, \textit{i.e.}, continuous or discrete measurements recorded over time, such as sensor readings, financial metrics, or physiological signals, their modalities often include: \textbf{1) Tabular:} Time-indexed records that are inherently organized in a tabular format, such as event logs, transaction records, or demographic information. \textbf{2) Text:} Time-stamped or domain-specific textual information --- like clinical notes, financial reports, news articles, or social media posts --- that provides contextual or interpretative insights.
\textbf{3) Image:} Visual data acquired as images over time, such as photographs, medical images (\textit{e.g.}, X-rays, MRI), satellite imagery, or visual representations generated from time series data. \textbf{4) Graph:} Relational data representing interactions or structural dependencies among entities that evolve. They are typically modeled as networks or graphs, where the connections may change dynamically. Although \textbf{audio} is widely studied as an independent modality in multi-modal research, we consider it a special form of time series in this survey and briefly discuss representative works within this scope.

\subsection{Common Datasets and Benchmarks}

Multi-modal time series datasets vary a lot and are domain-dependent, each with unique data characteristics and modalities. In Table \ref{tab:mm_datasets} we provide representative datasets categorized by domain, along with their respective modalities: 

\noindent\textbf{Healthcare}: In this domain, physiological signals (\textit{e.g.}, ECG, EEG) are extensively analyzed alongside textual data such as clinical notes, patient demographics, and tabular data including vital signs and laboratory results. Common datasets include MIMIC-III~\cite{Johnson2016}, a comprehensive dataset containing electronic health records (EHRs) of ICU patients with physiological measurements, clinical notes, and diagnostic information, widely used for tasks like patient monitoring, mortality prediction, and clinical decision support. MIMIC-IV~\cite{Johnson2021} is an extension of MIMIC-III, which provide detailed physiological signals, clinical narratives, medication records, and demographic data from a large population of critically ill patients, frequently utilized for predictive modeling, clinical outcome analysis, and health informatics research. Other notable healthcare datasets include ICBHI~\cite{Rocha2019}, which contains respiratory sound recordings paired with clinical annotations for respiratory disease classification; Coswara~\cite{Bhattacharya2023}, which provides respiratory audio samples and rich metadata for COVID-19 detection tasks; KAUH~\cite{KAUHdata}, which comprises audio records and corresponding annotations for healthcare analytics; PTB-XL~\cite{Wagner2020}, a large-scale ECG dataset annotated with diagnostic labels for cardiac monitoring and diagnosis; ZuCo~\cite{ZuCo2019, ZuCo2020}, which consists of simultaneous EEG and textual data from reading comprehension tasks, being useful for cognitive neuroscience studies; and Image-EEG~\cite{ImageEEG}, which pairs EEG signals with images of objects on a natural background, aiding studies in visual neuroscience and computer vision.

\smallskip

\noindent\textbf{Finance}: Datasets that combine time series data with financial news and reports are instrumental in financial analysis and modeling. Notable examples include ACL18~\cite{ACL18}, CIKM18~\cite{CIKM18}, and DOW30~\cite{chen2023chatgpt}. These datasets focus on high-trade-volume stocks from the U.S. stock markets, providing historical stock price data, such as opening, high, low, and closing prices; alongside related textual information, including tweets or financial news. Another large-scale dataset, FNSPID~\cite{FNSPID}, consists of stock prices and time-aligned financial news records, covering over 4,000 companies from 1999 to 2023.

\smallskip

\noindent\textbf{Multi-domain}: Datasets featuring general-purpose numerical time series combined with textual data are suitable for broad analytical applications. Examples include Time-MMD~\cite{liu2024timemmd}, which encompasses nine primary data domains: Agriculture, Climate, Economy, Energy, Environment, Health, Security, Social Good, and Traffic, while ensuring fine-grained alignment between time series and textual data; TimeCAP~\cite{lee2024timecap} compiles seven real-world time series datasets across three domains: weather, finance, and healthcare. To generate textual descriptions for each time series, a large language model (LLM) agent is employed, leveraging contextual information and domain-specific knowledge. NewsForecast~\cite{2024wangnewsforecast} integrates task-specific time series data with verified public news reports across various domains, including finance, energy, traffic, and cryptocurrency; TTC~\cite{kim2024multimodalforecasterjointlypredicting} is a meticulously curated, time-aligned dataset designed for multimodal forecasting. It consists of paired time series and text data synchronized to timestamps, spanning two distinct domains: climate science and healthcare; CiK~\cite{williams2024context} is a dataset comprising 71 forecasting tasks across seven real-world domains. Each task necessitates the integration of both numerical data and textual information. The covered domains include Climatology, Economics, Energy, Mechanics, Public Safety, Transportation, and Retail. The TSQA~\cite{kong2025timemqatimeseriesmultitask} dataset consists of 200k question-answer pairs derived from time series data across 12 domains: healthcare, finance, energy, traffic, environment, IoT, nature, transport, human activities, machine sensors, AIOps, and the web. These QA pairs are designed to support five key tasks: forecasting, imputation, anomaly detection, classification, and open-ended reasoning.

\vspace{0.2cm}

\noindent\textbf{Other domains}: Beyond the previously discussed major sectors, multi-modal time series analysis extends to various other domains. In \textit{Retail}, datasets such as VISUELLE~\cite{skenderi2024well} integrate numerical sales data with product images and textual descriptions, facilitating thorough analyses of consumer behavior and inventory management. The \textit{Internet of Things (IoT)} domain benefits from datasets such as LEMMA-RCA~\cite{zheng2024lemmarca}, which combine time series sensor data with textual metadata, enabling enhanced monitoring and more robust and secure methodologies that ensure the high performance of modern systems. In the \textit{Speech} domain, datasets like LRS3~\cite{2018LRS3} and VoxCeleb2~\cite{Chung_2018} integrate audio recordings with corresponding visual data, supporting advancements in speech recognition and speaker identification technologies. In the \textit{Traffic} domain, datasets like NYC-Taxi, NYC-Bike~\cite{Li2024UrbanGPTSL} contain spatial-temporal (ST) data alongside associated textual metadata. These integrations allow LLMs to effectively capture and utilize spatial-temporal contextual signals. In the \textit{Environment} domain, Terra~\cite{NEURIPS2024_Terra} collect 45 years of global geographic spatial-temporal data, supplemented with textual descriptions.

\section{Cross-modal Interactions with Time Series}
In this section, we conduct a detailed review of existing research on multi-modal time series analysis by thoroughly analyzing cross-modal interactions. We also elaborate how existing multi-modal methods are tailored for domain-specific applications in Section 5. The detailed taxonomy is provided in Table~\ref{tab:taxonomy}. 

\begin{figure}
    \centering
    \includegraphics[width=1\linewidth]{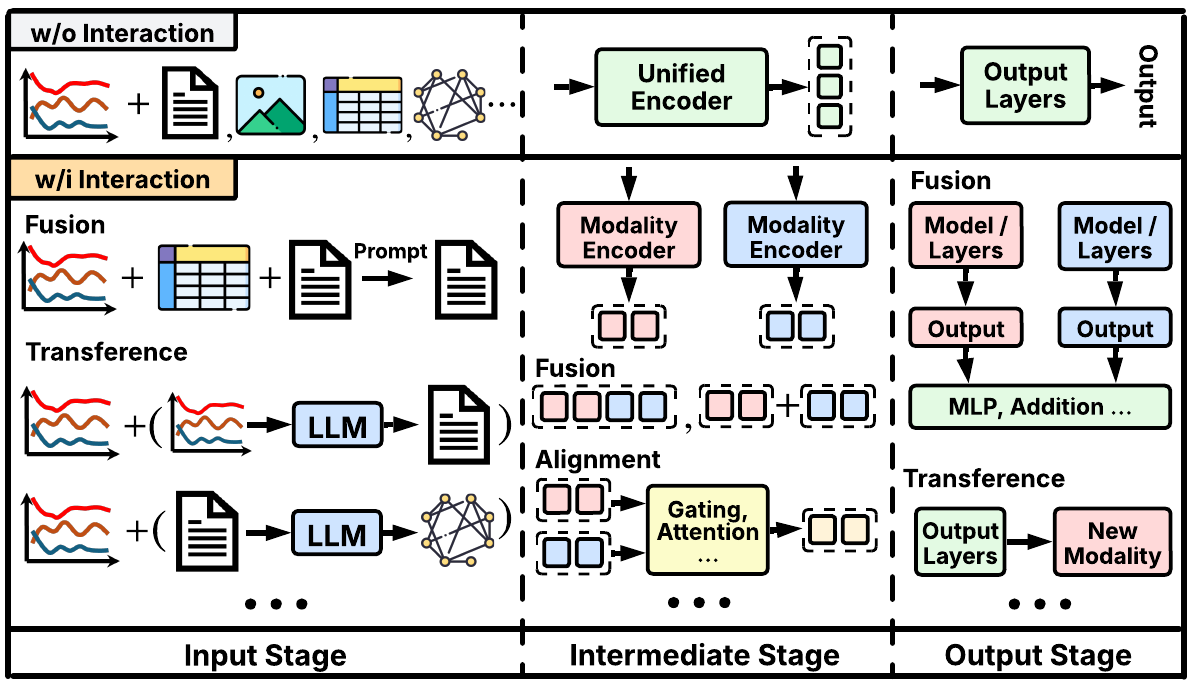}
    \caption{Categorization of cross-modal interaction methods and representative examples. }
    \label{fig:taxonomy-example}
    \vspace{-2mm}
\end{figure}

We define three fundamental types of interactions between time series and other modalities, including {\em fusion}, {\em alignment}, and {\em transference}, which occur at different stages within a framework --- {\em input}, {\em intermediate} (\textit{i.e.}, representations or intermediate outputs), and {\em output}.
The representative examples are provided in Figure~\ref{fig:taxonomy-example}.

\begin{table*}[t]
\label{tab:taxonomy}
  \setlength{\tabcolsep}{3.5pt}
  \renewcommand{\arraystretch}{0.61}
  \scriptsize
  \caption{\footnotesize{Taxonomy of representative multi-modal time series methods. \textit{Modality} refers to the different data modalities involved in each method. \textit{TS} represents standard time series, \textit{ST} denotes spatial time series. The \textit{Method} column lists the techniques used for each interaction, separated by semicolons, where each interaction may include one or more techniques, separated by commas. Superscripts in the \textit{Code} column include the URLs to Github repositories.}}
  
  \vspace{-0.25cm}
  \centering
  \begin{threeparttable}
 
     \begin{tabular}{lccccccccccl}
     \toprule
     \cmidrule{1-12}
     
    \multirow{2}{*}{Method} & \multirow{2}{*}{ Modality} & \multirow{2}{*}{Domain} & \multirow{2}{*}{Task} &\multicolumn{5}{c}{Cross-Modal Interaction} & \multirow{2}{*}{Large Model} & \multirow{2}{*}{Year} & \multirow{2}{*}{Code}  \\
    \cmidrule(lr){5-9}
    
     &   &  & & Stage & Fusion & Align. & Trans. & Method   \\
    \midrule
    \cmidrule{1-12}

     Time-MMD~\cite{liu2024timemmd} & TS, Text & General & Forecasting & Output & \cmark & \xmark & \xmark & Addition
     & Multiple & 2024 & Yes\textsuperscript{\href{https://github.com/AdityaLab/Time-MMD}{[1]}}\\


     \midrule
     \multirow{2}{*}{Wang et al.~\cite{2024wangnewsforecast}}
     & \multirow{2}{*}{TS, Text} & \multirow{2}{*}{General} & \multirow{2}{*}{Forecasting} & Input & \cmark & \xmark & \xmark & Prompt & LLaMa2 & \multirow{2}{*}{2024} & \multirow{2}{*}{Yes\textsuperscript{\href{https://github.com/ameliawong1996/From_News_to_Forecast}{[2]}}} \\
     & &&& Intermediate & \cmark & \cmark & \xmark & Prompt; LLM Reasoning & GPT-4 Turbo & & \\

 
     \midrule
     GPT4MTS~\cite{gpt4mts} & TS, Text & General & Forecasting & Intermediate & \cmark & \cmark & \xmark & Addition; Self-attention  & GPT-2 & 2024 & No \\

     \midrule
     \multirow{2}{*}{TimeCMA~\cite{liu2024timecma}} &\multirow{2}{*}{TS, Text} & \multirow{2}{*}{General} & \multirow{2}{*}{Forecasting} & Input &\xmark &\xmark & \cmark & Meta-description & \multirow{2}{*}{GPT-2} & \multirow{2}{*}{2025} & \multirow{2}{*}{Yes\textsuperscript{\href{https://github.com/ChenxiLiu-HNU/TimeCMA}{[3]}}}  \\ 
     &&&&Intermediate &\cmark &\cmark & \xmark & Addition; Cross-attention\\

    \midrule
    \multirow{2}{*}{MOAT~\cite{lee2024moat}} & \multirow{2}{*}{TS, Text} &  \multirow{2}{*}{General} & \multirow{2}{*}{Forecasting} & {Intermediate} & \cmark & \cmark & \xmark & {Concat.; Self-attention} & \multirow{2}{*}{S-Bert} & \multirow{2}{*}{2024} & \multirow{2}{*}{No} \\
       &   &   &  & Output & \cmark & \xmark & \xmark & Offline Synthesis (MLP) &\\

    \midrule
     \multirow{3}{*}{TimeCAP~\cite{lee2024timecap}} & \multirow{3}{*}{TS, Text} & \multirow{3}{*}{General} & \multirow{3}{*}{Classification} & Input & \xmark & \xmark & \cmark & LLM Generation & \multirow{3}{*
     }{Bert, GPT-4} & \multirow{3}{*
     }{2024} & \multirow{3}{*}{No}\\
     &&&& Intermediate & \cmark & \cmark & \xmark & Concat.; Self-attention, Retrieval & & & \\
     &&&& Output & \cmark & \xmark & \xmark & Addition & & & \\

    \midrule

    \multirow{2}{*}{TimeXL~\cite{jiang2025explainablemultimodaltimeseries}} & \multirow{2}{*}{TS, Text} & \multirow{2}{*}{General} & {Classification} & Intermediate & \cmark & \cmark & \xmark &  Concat., Prompt; LLM Reasoning & {Bert, S-Bert} & \multirow{2}{*}{2025} & \multirow{2}{*}{No}\\ 
    &   &   & Forecasting & Output & \cmark & \xmark & \xmark & Addition & GPT-4o\\

     \midrule
     Hybrid-MMF~\cite{kim2024multimodalforecasterjointlypredicting} & TS, Text & General & Forecasting & Intermediate & \cmark  & \xmark & \xmark & Concat. & {GPT-4o} & 2024 &Yes\textsuperscript{\href{https://github.com/Rose-STL-Lab/Multimodal\_Forecasting}{[4]}} \\ 

     \midrule
     \multirow{2}{*}{Time-LLM~\cite{jin2024timellm}} & \multirow{2}{*}{TS, Text} & \multirow{2}{*}{General} & \multirow{2}{*}{Forecasting} & Input & \xmark & \xmark & \cmark &  Meta-description & \multirow{2}{*}{LLaMA, GPT-2} & \multirow{2}{*}{2024} & \multirow{2}{*}{Yes\textsuperscript{\href{https://github.com/kimmeen/time-llm}{[5]}}}\\ 
    &   &   &  & Intermediate & \cmark & \cmark & \xmark & Concat.; Self-attention &\\
    
    \midrule
    \multirow{2}{*}{Time-VLM~\cite{zhong2025timevlm}}
   & \multirow{2}{*}{TS, Text, Image}
   & \multirow{2}{*}{General} 
   & \multirow{2}{*}{Forecasting}
   & Input 
   & \xmark 
   & \xmark 
   & \cmark 
   & Feat. Imaging, Meta-description 
   & {ViLT, CLIP}
   & \multirow{2}{*}{2025}
   & \multirow{2}{*}{No}  \\
   &   &   &  & Intermediate & \cmark & \cmark & \xmark & Addition; Gating, Cross-attention & BLIP-2\\
     
     \midrule

     \multirow{2}{*}{Unitime~\cite{unitime}} & \multirow{2}{*}{TS, Text} & \multirow{2}{*}{General} & \multirow{2}{*}{Forecasting} & Input & \xmark & \xmark & \cmark &  Meta-description & \multirow{2}{*}{GPT-2} & \multirow{2}{*}{2024} & \multirow{2}{*}{Yes\textsuperscript{\href{https://github.com/liuxu77/UniTime}{[6]}}}\\ 
    &   &   &  & Intermediate & \cmark & \cmark & \xmark & Concat.; Self-attention &\\

    \midrule
    
    TESSA~\cite{lin2024decodingtimeseriesllms} & TS, Text & General & Annotation & Intermediate & \cmark & \cmark & \cmark & Prompt; RL; LLM Generation & GPT-4o & 2024 & No\\

   \midrule
   InstrucTime~\cite{InstrucTime}
   & TS, Text
   & General
   & Classification
   & Intermediate
   & \cmark 
   & \cmark 
   & \xmark 
   & Concat.; Self-attention
   & GPT-2
   & 2025
   & Yes\textsuperscript{\href{https://github.com/Mingyue-Cheng/InstructTime}{[7]}}  \\ 

    \midrule

    MATMCD~\cite{shen2024exploringmultimodalintegrationtoolaugmented} & {TS, Text, Graph} & {General} 
    & Causal Discovery & Intermediate & \cmark & \cmark &\cmark & Prompt; LLM Reasoning; Supervision & Multiple & 2025 & No\\

    \midrule

    STG-LLM~\cite{liu2024can} & ST, Text & General & Forecasting & Intermediate & \cmark & \cmark & \xmark & Concat.; Self-attention & GPT-2 & 2024 & No\\ 

    \midrule
     \multirow{2}{*}{TableTime~\cite{wang2025tabletimereformulatingtimeseries}}
     & \multirow{2}{*}{TS, Text} 
     & \multirow{2}{*}{General}
     & \multirow{2}{*}{Classification}
     & \multirow{2}{*}{Input}
     & \multirow{2}{*}{\cmark} & \multirow{2}{*}{\xmark} & \multirow{2}{*}{\cmark}
     & \multirow{2}{*}{Prompt; Reformulate}
     & \multirow{2}{*}{Multiple}
     & \multirow{2}{*}{2024}
     & \multirow{2}{*}{Yes\textsuperscript{\href{https://anonymous.4open.science/r/TableTime-5E4D/README.md}{[8]}}}
     \\
     &&&&&&&&&\\

    \midrule
    ContextFormer~\cite{chattopadhyay2025contextmattersleveragingcontextual} & TS, Tabular & General &Forecasting &Intermediate & \cmark & \cmark &\xmark & Addition; Cross-attention & No & 2025 & No \\

    \midrule
    \multirow{2}{*}{Time-MQA~\cite{kong2025timemqatimeseriesmultitask}}
    & \multirow{2}{*}{TS, Text}
    & \multirow{2}{*}{General}
    & \multirow{2}{*}{Multiple}
    & \multirow{2}{*}{Input}
    & \multirow{2}{*}{\cmark} & \multirow{2}{*}{\xmark} & \multirow{2}{*}{\xmark}
    & \multirow{2}{*}{Prompt}
    & \multirow{2}{*}{Multiple}
    & \multirow{2}{*}{2025}
    & \multirow{2}{*}{Yes\textsuperscript{\href{https://huggingface.co/Time-QA}{[9]}}}
    \\
    &&&&&&&&&

    \\

    \midrule

    MAN-SF~\cite{sawhney-etal-2020-deep} & TS, Text, Graph & Finance & Classification & Intermediate & \cmark & \cmark & \xmark & Bilinear; Graph Convolution & USE & 2020 & No\\
    
    \midrule

    \multirow{2}{*}{Bamford et al.~\cite{latentspaceprojection}} & TS, Text & \multirow{2}{*}{Finance} & \multirow{2}{*}{Retrieval} &  {Intermediate} &  {\xmark} &  {\cmark} & \xmark & \multirow{2}{*}{Supervision} & \multirow{2}{*}{S-Bert} & \multirow{2}{*}{2024} & \multirow{2}{*}{No}\\

   & TS, Image &  &  & Output & \xmark & \xmark & \cmark & &  \\

    \midrule

    \multirow{2}{*}{Chen et al.~\cite{chen2023chatgpt}} & \multirow{2}{*}{TS, Text, Graph} & \multirow{2}{*}{Finance} & \multirow{2}{*}{Classification} & \multirow{2}{*}{Intermediate} & \xmark & \xmark & \cmark & LLM Generation & \multirow{2}{*}{ChatGPT} & \multirow{2}{*}{2023} & \multirow{2}{*}{No} \\ 
    &&&&& \cmark & \cmark & \xmark & Concat.; Graph Convolution\\

    \midrule

    Xie et al.~\cite{xie2023wall} & TS, Text & Finance & Classification & Input & \cmark & \xmark & \xmark & Prompt & ChatGPT & 2023 & No\\ 

     \midrule
     Yu et al.~\cite{2023-yu-financeforecasting}
     & TS, Text
     & Finance
     & Forecasting
     & Input
     & \cmark & \xmark & \xmark
     & Prompt
     & GPT-4, Open LLaMA
     & 2023
     & No
     \\

    \midrule
    MedTsLLM~\cite{chan2024medtsllm} & TS, Text, Tabular & Healthcare & Multiple & Intermediate & \cmark & \cmark &\xmark & Concat.; Self-attention & Llama2 & 2024 & Yes\textsuperscript{\href{https://github.com/flixpar/med-ts-llm}{[10]}} \\

    





     \midrule
     RespLLM~\cite{RespLLM} 
     & TS (Audio), Text 
     & Healthcare 
     & Classification 
     & Intermediate 
     & \cmark & \cmark & \xmark 
     & Addition, Self-attention 
     & OpenBioLLM-8B 
     & 2024 
     & No \\

     \midrule
     METS~\cite{li2023frozen} 
     & TS, Text
     & Healthcare
     & Classification
     & Output
     & \xmark & \cmark & \xmark
     & Contrastive 
     & ClinicalBert
     & 2023
     & No \\

     \midrule
     Wang et al.~\cite{Wang2021OpenVE}
     & TS, Text
     & Healthcare
     & Classification
     & Intermediate
     & \xmark & \xmark & \cmark
     & Supervision
     & Bart, Bert, RoBerta
     & 2021
     & No \\

     \midrule
     EEG2TEXT~\cite{2024EEG2Text}
     & TS, Text
     & Healthcare
     & Generation
     & Output
     & \xmark & \xmark & \cmark
     & Self-supervision, Supervision
     & Bart
     & 2024
     & No \\

     \midrule
     MEDHMP~\cite{2023MEDHMP}
     & TS, Text 
     & Healthcare
     & Classification
     & Intermediate
     & \cmark & \cmark & \xmark
     & Concat.; Self-attention, Contrastive
     & ClinicalT5
     & 2023
     & Yes\textsuperscript{\href{https://github.com/XiaochenWang-PSU/MedHMP}{[11]}}
     \\


%
    \midrule
     Deznabi et al.~\cite{deznabi-etal-2021-predicting} & TS, Text & Healthcare & Classification & Intermediate  & \cmark & \xmark & \xmark & Concat. & Bio+Clinical Bert & 2021 & Yes \textsuperscript{\href{https://github.com/Information-Fusion-Lab-Umass/ClinicalNotes_TimeSeries}{[12]}}   \\



    \midrule

     Niu et al.~\cite{niu2023deep} & TS, Text & Healthcare & Classification & Intermediate & \cmark & \cmark & \xmark &  Concat.; Cross-attention & BioBERT & 2023 &  No\\

    \midrule
     Yang et al.~\cite{yang2021how} & TS, Text & Healthcare 
     & Classification
     & Intermediate & \cmark  & \cmark & \xmark & Concat., Addition; Gating & ClinicalBERT & 2021 & Yes\textsuperscript{\href{https://github.com/emnlp-mimic/mimic}{[13]}}  \\


    \midrule
     \multirow{2}{*}{Liu et al.~\cite{liu2023large}} & \multirow{2}{*}{TS, Text} & \multirow{2}{*}{Healthcare} & Classification & \multirow{2}{*}{Input} & \multirow{2}{*}{\cmark} & \multirow{2}{*}{\xmark} & \multirow{2}{*}{\xmark} & \multirow{2}{*}{Prompt} & \multirow{2}{*}{PaLM} & \multirow{2}{*}{2023}& \multirow{2}{*}{Yes\textsuperscript{\href{https://github.com/marianux/ecg-kit}{[14]}}}   \\

     &&&Regression\\

     


     


    \midrule

    xTP-LLM~\cite{guo2024towards} & ST, Text & Traffic & Forecasting & Input & \cmark & \xmark & \cmark & Prompt; Meta-description & Llama2-7B-chat & 2024 &  Yes\textsuperscript{\href{https://github.com/Guoxs/xTP-LLM}{[15]}} \\

    \midrule
    UrbanGPT~\cite{Li2024UrbanGPTSL}
    & ST, Text
    & Traffic
    & Forecasting
    & Input
    & \cmark
    & \xmark
    & \cmark
    & Prompt; Meta-description
    & Vicuna-7B
    & 2024
    & Yes\textsuperscript{\href{https://github.com/HKUDS/UrbanGPT}{[16]}}\\

    \midrule
    CityGPT~\cite{feng2024citygpt} & ST, Text & Mobility & Multiple & Input & \cmark & \xmark & \xmark & Prompt & Multiple & 2025 & Yes\textsuperscript{\href{https://github.com/tsinghua-fib-lab/CityGPT}{[17]}}\\

    \midrule


    MULAN~\cite{zheng2024mulan} & {TS, Text, Graph} & {IoT} 
    & Causal Discovery
    & Intermediate & \cmark & \cmark & \cmark & Addition; Contrastive; Supervision & No & 2024 & No \\

    \midrule

    MIA~\cite{XING2023108567} & TS, Image & IoT & Anomaly Detection & Intermediate & \cmark & \cmark & \xmark & Addition; Cross-attention, Gating & No & 2023 & No\\

    \midrule

    Ekambaram et al.~\cite{ekambaram2020attention}
    & TS, Image, Text 
    & Retail
    & Forecasting 
    & Intermediate 
    & \cmark 
    & \cmark
    & \xmark  
    & Concat.; Self \& Cross-attention
    & No
    & 2020 
    & Yes\textsuperscript{\href{https://github.com/HumaticsLAB/AttentionBasedMultiModalRNN}{[18]}} \\
    \midrule

    Skenderi et al.~\cite{skenderi2024well}
    & TS, Image, Text 
    & Retail
    & Forecasting 
    & Intermediate 
    & \cmark 
    & \cmark 
    & \xmark  
    & Concat.; Cross-attention
    & No
    & 2024 
    & Yes\textsuperscript{\href{https://github.com/HumaticsLAB/GTM-Transformer}{[19]}} \\
    \midrule


    VIMTS~\cite{zhao2022vimts} & ST, Image & Environment & Imputation & Intermediate & \cmark & \cmark & \xmark & Concat.; Supervision & No & 2022 & No\\
    \midrule

    LITE~\cite{li2024lite} 
    & ST, Text, Image 
    & Environment
    & Forecasting
    & Intermediate 
    & \cmark 
    & \cmark 
    & \xmark
    & Concat.; Self-attention 
    & LLaMA-2-7B
    & 2024
    & Yes\textsuperscript{\href{https://github.com/hrlics/LITE}{[20]}}  \\

    \midrule
     AV-HuBERT~\cite{shi2022learning}
     & TS (Audio), Image
     & Speech
     & Classification
     & Intermediate
     & \cmark & \cmark & \xmark
     & Concat.; Self-attention
     & HuBert
     & 2022
     & Yes\textsuperscript{\href{https://github.com/facebookresearch/av_hubert}{[21]}}
     \\
    
    \midrule
    SpeechGPT ~\cite{zhang2023speechgpt}
    & TS(Audio), Text
    & Speech
    & Generation
    & intermediate
    & \cmark
    & \cmark
    & \xmark
    & Concat.; Self-attention
    & LLaMA-13B
    & 2023
    & Yes\textsuperscript{\href{https://github.com/0nutation/SpeechGPT}{[22]}}\\

     \midrule
     LA-GCN~\cite{lagcn} & ST, Text & Vision & Classification & Intermediate & \xmark & \cmark & \xmark & Supervision & Bert & 2023 & Yes\textsuperscript{\href{https://github.com/damNull/LAGCN}{[23]}}\\ 

    \bottomrule
    \cmidrule{1-12}
    \end{tabular}

    \end{threeparttable}
 \vspace{-0.4cm}
 \label{tab:taxonomy}
\end{table*}

\subsection{Fusion}
\textbf{Fusion} refers to the process of integrating heterogeneous modalities in a way that captures complementary information across diverse sources to improve time series modeling.
To fuse multi-modal inputs, a common practice is to directly integrate time series, tabular data and texts into a unified textual prompt, then use it to query LLMs for downstream tasks. This is typically facilitated by instruction fine-tuning for task-oriented analysis~\cite{2024wangnewsforecast,kong2025timemqatimeseriesmultitask,liu2023large,guo2024towards,Li2024UrbanGPTSL,feng2024citygpt,zhang2023speechgpt}. Some works also leverage the zero-shot reasoning and inference capability of pretrained LLMs (\textit{e.g.}, GPT-4 and its variants)~\cite{wang2025tabletimereformulatingtimeseries,xie2023wall,2023-yu-financeforecasting}. Recent research efforts like TaTS~\cite{li2025languageflowtimetimeseriespaired} attempt to integrate paired text embedding as an additional variable of time series for temporal modeling, yielding competitive task performance.

Most existing methods perform cross-modal fusion at the intermediate stage, such as adding and concatenating multi-modal representations, where each individual modal encoder first maps the raw data into a shared latent space. Addition combines time series and other modalities by summing up encoded representations, effectively blending shared information while preserving their interconnections in the latent space~\cite{gpt4mts,liu2024timecma,chattopadhyay2025contextmattersleveragingcontextual,zheng2024mulan,RespLLM,yang2021how}. On the other hand, concatenation stacks multi-modal representations along the same dimension, retaining modality-specific characteristics and allowing models to capture joint relationships between the modalities~\cite{kim2024multimodalforecasterjointlypredicting,deznabi-etal-2021-predicting,khadanga2019using,khadanga2019using}. To effectively leverage cross-modal information, existing methods often incorporate alignment designs after concatenating representations~\cite{lee2024moat,lee2024timecap,jiang2025explainablemultimodaltimeseries,jin2024timellm,unitime,InstrucTime,liu2024can,chan2024medtsllm,chen2023chatgpt,2023MEDHMP,niu2023deep,yang2021how,ekambaram2020attention,skenderi2024well,zhao2022vimts,li2024lite,shi2022learning}. Alignment is also used in the aforementioned additions, which will be detailed in Section \ref{sec:alignment}.  

When fusion is performed at the output level, different modalities contribute separately to the final output, allowing each modality to retain its unique predictive signal~\cite{liu2024timemmd,lee2024moat,lee2024timecap,jiang2025explainablemultimodaltimeseries}. Time-MMD~\cite{liu2024timemmd} provides a paradigm that fuses predictions from both state-of-the-art forecasters and a pretrained language model with a projection layer, in an end-to-end manner. MOAT~\cite{lee2024moat} introduces a two-stage framework for multi-modal time series forecasting. In the first stage, the model is optimized to generate forecasts from decomposed time series and text embeddings. In the second stage, an offline synthesis via MLP is applied to dynamically fuse different components, yielding the final forecast based on their relative contributions. Beyond fusing outputs from a single model, TimeCAP~\cite{lee2024timecap} enhances performance by combining predictions from both a multi-modal predictor and a pretrained LLM, which synergizes the gradient-based method and LLM agents reasoning on real-world contexts. Output fusion gains advantage of design flexibility and robustness, but it may not fully utilize the complementary relationship between modalities without additional countermeasures.  

Cross-modal fusion relies on well-aligned multi-modal data for effective exploitation of the contextual information. However, ideally-aligned data may not be given in real-world scenarios. As such, existing methods also leverage alignment mechanisms to mitigate the challenge.

\subsection{Alignment}\label{sec:alignment}
\textbf{Alignment} ensures that the relationships between different modalities are preserved and semantically coherent when integrated into a unified learning framework. At the input level, we primarily refer alignment to data preprocessing techniques that aim at mitigating temporal misalignment caused by missing values, irregular sampling intervals, and differing granularities across modalities. This process is crucial for ensuring that data from multiple sources are properly synchronized before fusion, where domain knowledge is usually needed to handle such inconsistencies~\cite{liu2024timemmd,2024wangnewsforecast,NEURIPS2024_Terra}. In addition, none of the existing methods we reviewed explicitly perform output alignment. However, the aforementioned output fusion can be easily adapted to alignment through the incorporation of a gating or attention mechanism that we will introduce shortly. 

Alignment at the intermediate stage plays a crucial role in multi-modal interactions. We first introduce the alignment of multi-modal representations, spanning a range of techniques from model component design to learning objectives. The common component designs include self-attention~\cite{gpt4mts,lee2024moat,lee2024timecap,jin2024timellm,unitime,InstrucTime,liu2024can,chan2024medtsllm,li2024lite,shi2022learning}, cross-attention~\cite{liu2024timecma,zhong2025timevlm,chattopadhyay2025contextmattersleveragingcontextual,yang2021how,niu2023deep,XING2023108567,ekambaram2020attention,skenderi2024well} and gating mechanisms~\cite{zhong2025timevlm,XING2023108567}. 
Self-attention is often used to fuse multi-modal representations. It enables a joint and undirected alignment across all modalities by dynamically attending to important features. 
Given multi-modal embeddings $E_{\mathrm{mm}} \in \mathbb{R}^{n \times d}$, where $n$ is the total number of modality tokens and $d$ is the embedding dimension, self-attention is computed as follows:
$$
\operatorname{ Attention }(E_{\mathrm{mm}})=\text { softmax }\left(\frac{Q K^{\top}}{\sqrt{d_k}}\right) V
$$
where the queries $Q$, keys $K$, and values $V$ are linear projections of $E_{\mathrm{mm}}$: $Q=E_{\mathrm{mm}} W_Q$, $K=E_{\mathrm{mm}} W_K$, $V=E_{\mathrm{mm}} W_V$ with learnable weights of dimensionality $d_k$: $W_Q, W_K, W_V \in \mathbb{R}^{d \times d_k}$.

In cross-attention, time series serves as the 
query modality to get contextualized by other modalities, providing a directed alignment that ensure auxiliary modalities contribute relevant contextual information while preserving the temporal structure of time series. Given a query embedding $E_{\mathrm{ts}} \in \mathbb{R}^{n \times d}$ and that of auxiliary modalities $E_{c} \in \mathbb{R}^{n \times d}$ as keys and values:
$$
\operatorname{CrossAttention}\left(E_{\mathrm{ts}}, E_{\mathrm{c}}\right)=\operatorname{softmax}\left(\frac{Q_{\mathrm{ts}} K_{\mathrm{c}}^{\top}}{\sqrt{d_k}}\right) V_{\mathrm{c}}
$$
where the query, key and value are denoted as $Q_{\mathrm{ts}}=E_{\mathrm{ts}} W_Q$, $K_{\mathrm{c}}=E_{\mathrm{c}} W_K$, $V_{\mathrm{c}}=E_{\mathrm{c}} W_V$. Note that existing methods adopt multi-head attentions, which is omitted here for simplicity.

Similarly, the gating mechanism is a parametric filtering operation that explicitly regulates the influence of time series and other modalities on the fused embeddings in $E$:
$$
G = \sigma(W_g [E_{\mathrm{ts}}; E_c] + b_g), \quad E = G \odot E_{\mathrm{ts}} + (1 - G) \odot E_c
$$
where $\sigma(\cdot)$ denotes the sigmoid function, the learnable weight and bias are denoted as $W_g \in \mathbb{R}^{2d \times d}$ and $b_g \in \mathbb{R}^{d}$, respectively.

When a graph modality is available from external contexts, the underlying topological insights can be leveraged for graph-based alignment~\cite{chen2023chatgpt,sawhney-etal-2020-deep}. Unlike the above methods that rely solely on feature interactions, it explicitly aligns multi-modal representations with relational structures through graph convolution, enabling context-aware feature propagation across modalities.

Representation alignments can also be achieved by learning objectives~\cite{zheng2024mulan,li2023frozen,latentspaceprojection,zhao2022vimts,lagcn}. For example, MULAN~\cite{zheng2024mulan} extracts modality-invariant and modality-specific representations from multi-modal time series. It employs contrastive learning to enhance cross-modal alignment by maximizing the similarity between invariant representations across modalities while minimizing the similarity between invariant and specific representations of the same modality. Moreover, Bamford \textit{et al.}~\cite{latentspaceprojection} align cross-modal representations by using the mean of uni-modal cosine similarities as the target similarity and optimizing cross-modal similarity via cross-entropy loss, which effectively connects both modalities in a shared latent space for time series retrieval tasks. In general, this branch of methods is effective as it directly integrates the alignment objective into the optimization process, ensuring that meaningful representations are explicitly learned.

Lastly, we introduce the intermediate alignment of component outputs within a framework, extending beyond representation alignment within a model. 
The most recent studies explore the synergy between time series models and LLM agents, leveraging the strong reasoning capabilities of pretrained LLMs to provide contextual understanding and calibration in real-world scenarios~\cite{lee2024timecap,jiang2025explainablemultimodaltimeseries,2024wangnewsforecast,shen2024exploringmultimodalintegrationtoolaugmented,lin2024decodingtimeseriesllms}. 
We briefly discuss a few representative examples for demonstration. TimeCAP~\cite{lee2024timecap} utilizes the embedding space of a trained multi-modal encoder to retrieve in-context examples with the highest cosine similarity. These retrieved examples with ground truth labels are then fed, along with the query text, into an LLM to provide contextual guidance and improve outcome prediction. 
TimeXL~\cite{jiang2025explainablemultimodaltimeseries} incorporates a multi-modal prototype-based encoder to generate explainable case-based rationales for both time series and texts, integrating three LLM agents, where prediction, reflection, and refinement LLMs collaborate to iteratively enhance prediction accuracy, identify textual inconsistencies or noise, and calibrate textual contexts, yielding more accurate predictions and explanations. NewsForecast~\cite{2024wangnewsforecast} also employs reflection in language agents to iteratively select relevant news from a large database, enhancing alignment of textual information for text-based forecasting.
Similarly, MATMCD~\cite{shen2024exploringmultimodalintegrationtoolaugmented} ensures alignment between statistical causal discovery on time series and LLM reasoning on textual context by leveraging iterative self-reflective tool-calling to structure textual context, which is then used to explain and refine causal constraints.

In a nutshell, alignment aims to calibrate real-world contexts and effectively capture relevant multi-modal elements for a semantically coherent time series modeling. It enhances task performance, robustness and explanation, ensuring that models leverage meaningful contextual information for improved decision-making.

\subsection{Transference}
\textbf{Transference} refers to the process of mapping between different modalities. It allows one modality to be inferred, translated, or synthesized from another. This concept plays a crucial role across different stages of multi-modal time series analysis. The input-level transference typically serves for modality augmentation. It helps introduce contextual priors, enrich training samples, and provide alternative representations. This is particularly useful in scenarios of data scarcity and imbalance. In existing literature, a common practice is to use meta information to describe the narrative of real-world contexts (\textit{e.g.}, domain, data statistics and granularity, variable descriptions, other co-variates, \textit{etc.})~\cite{liu2024timecma,jin2024timellm,zhong2025timevlm,unitime,wang2025tabletimereformulatingtimeseries,guo2024towards,Li2024UrbanGPTSL,feng2024citygpt} or leverage pretrained LLMs to generate fine-grained textual contexts~\cite{lee2024timecap} or graphs~\cite{chen2023chatgpt} for real-world time series, serving as an augmented modality. In addition to texts, time series can also be transformed into high-dimensional images via feature imaging, such as stacking the original data with frequency and periodicity features~\cite{zhong2025timevlm}. Alternatively, time series can be represented in tabular form, transforming time series analysis into a table understanding task~\cite{wang2025tabletimereformulatingtimeseries}. Note that the aforementioned uni-modal methods for transforming time series into other single modalities can also be integrated into multi-modal time series frameworks~\cite{li2023time,prithyani2024feasibility,daswani2024plots,zhou2024can,zhuang2024see,hoo2025tabular}.
The exploitation of input-level transference is two-fold. First, the embedding of generated modality can serve as semantic anchors that guides time series modeling via representation alignment, improving downstream supervised tasks~\cite{liu2024timecma,jin2024timellm,zhong2025timevlm,unitime}. Second, it provides additional contextual guidance for pretrained LLMs through input fusion and prompting ~\cite{wang2025tabletimereformulatingtimeseries,guo2024towards,Li2024UrbanGPTSL,feng2024citygpt}.

At the intermediate~\cite{lin2024decodingtimeseriesllms,zheng2024mulan,shen2024exploringmultimodalintegrationtoolaugmented,chen2023chatgpt,Wang2021OpenVE} and output~\cite{latentspaceprojection,2024EEG2Text} levels, transference are more task-oriented. 
The output-level transference typically refers to the end-to-end generation of new modalities, such as text-based and image-based time series retrieval, where users provide textual descriptions or sketched trends to query relevant time series data~\cite{latentspaceprojection}. This also includes EEG-to-text conversion, enabling direct transformation from physiological signals to human-readable narratives~\cite{2024EEG2Text}.

The output of intermediate transference typically serves as an initial solution to be refined for modality generation tasks~\cite{lin2024decodingtimeseriesllms,zheng2024mulan,shen2024exploringmultimodalintegrationtoolaugmented} or a medium to be inferred for predictive tasks~\cite{Wang2021OpenVE}, facilitating downstream reasoning and further alignment within the multi-modal framework. MATMCD~\cite{shen2024exploringmultimodalintegrationtoolaugmented} generates an initial causal graph from time series, achieving modality transference in the intermediate level. Subsequently, it incorporates textual modality to refine the causal graph, ensuring improved alignment and interpretability. Moreover, Wang~\textit{et al.}~\cite{Wang2021OpenVE} adopt a two-stage mechanism for sentiment classification based on EEG data, where the model first converts EEG signals into reading texts and then employs a pretrained LLM based on texts for classification, achieving impressive zero-shot results.

\section{Applications of Multi-modal Time Series Analysis}
In this section, we review the existing applications of multi-modal time series analysis for both standard and spatial time series, covering diverse domains such as healthcare, finance, transportation, environment, retail, and the Internet of Things (IoT).

\subsection*{Standard Time Series}

\subsection{Healthcare}

Recent studies in healthcare highlight the multi-modal analysis of diverse medical data sources, such as EHRs (Electronic Health Records, containing lab values and clinical reports, \textit{etc.}), audio, EEG (Electroencephalogram), ECG (Electrocardiogram), and other wearable and medical sensor recordings,
for better disease diagnosis and patient monitoring. 
For multi-modal analysis on EHR data, a common modeling strategy involves the interaction between lab values and clinical reports, including the concatenation~\cite{deznabi-etal-2021-predicting} and attention mechanisms~\cite{yang2021how,niu2023deep} on modality embeddings. 
Moreover, existing methods explore different modeling techniques to better exploit the clinical notes, via domain-specific text encoders (\textit{e.g.}, ClinicalBERT~\cite{yang2021how,huang2020clinicalbert} and BioBERT~\cite{niu2023deep,biobert}) and different processing strategies. For example, text embeddings can be modeled separately based on patient groups~\cite{yang2021multimodal} or through a decaying mechanism based on time intervals~\cite{khadanga2019using} before interacting with time series embeddings, which leads to improved mortality prediction. 

In addition to EHRs, multi-modal modeling methods have been tailored for audio~\cite{RespLLM}, ECG~\cite{li2023frozen}, and EEG~\cite{Wang2021OpenVE}.
Zhang \textit{et al.}~\cite{RespLLM} focus on a respiratory health classification task by integrating both audio and textual descriptions. 
Li \textit{et al.}~\cite{li2023frozen} propose a multi-modal contrastive learning framework, constructing positive and negative samples by pairing patients' report texts with corresponding ECG signals for self-supervised pretraining. The classification task is then performed by computing the cosine similarity between different text representations and the target ECG representation. 
Wang \textit{et al.}~\cite{Wang2021OpenVE} propose a two-stage method for zero-shot EEG-based sentiment classification. First, a pretrained BART model is used for EEG-to-text decoding, followed by a trained text sentiment classifier that converts the generated text into sentiment categories.

Similarly, Liu \textit{et al.} ~\cite{liu2023large} fuses physiological and behavioral time-series sensor data with real-world contextual information to effectively harness LLMs for wellness assessment. By fine-tuning the models with few-shot question-answer pairs that include contextual details, they improve performance on various healthcare tasks—such as cardiac signal analysis, physical activity recognition, and calorie-burn estimation, which outperform both supervised feedforward neural networks and zero-shot LLM baselines.

\subsection{Finance}
Recently, multi-modal time series analysis has received increasing attention in financial applications. Yu \textit{et al.}~\cite{2023-yu-financeforecasting} and Xie \textit{et al.}~\cite{xie2023wall} focus on stock prediction by integrating stock price movements, company profiles, and news directly into structured LLM prompts, enabling models to perform reasoning over multiple modalities. Yu \textit{et al.}~\cite{2023-yu-financeforecasting} applies GPT-4 and Open LLaMA to forecast NASDAQ-100 stock returns through instruction-based prompting and fine-tuning, demonstrating that structured LLM-driven inference can outperform traditional econometric models. Meanwhile, Xie \textit{et al.}~\cite{xie2023wall} conducts a zero-shot analysis of ChatGPT's capabilities for multi-modal stock movement prediction, incorporating CoT prompting to assess the impact of social media sentiment on stock trends. Chen \textit{et al.}\cite{chen2023chatgpt} and Sawhney\textit{et al.}\cite{sawhney-etal-2020-deep} also incorporate graph structures for stock movement prediction. For instance, Chen \textit{et al.}\cite{chen2023chatgpt} uses ChatGPT to infer dynamic stock relationship graphs from news, which reflects market conditions and enhances prediction accuracy.

Beyond the predictive tasks, Bamford \textit{et al.} ~\cite{latentspaceprojection} proposes a multi-modal retrieval framework, where the model aligns both modalities in a shared latent space through contrastive learning. This framework allows users to search for financial time series through textual descriptions or sketched trends, offering greater flexibility. It also significantly improves retrieval speed and accuracy compared to traditional SQL-based search methods.

\subsection{Others}

Multi-modal time series analysis also exists in other domains, such as retail, IoT, computer vision and audio. In the retail 
sector, Ekambaram {\em et al.}~\cite{ekambaram2020attention} utilizes product images and textual descriptions, including attributes like color, pattern, and sleeve style, while incorporating temporal and exogenous features for new product sales forecasting. More recently, Skenderi {\em et al.}~\cite{skenderi2024well} 
integrates additional modality data, including product images and text descriptions, along with Google Trends data for sales forecasting. In IoT applications, MIA~\cite{XING2023108567} enhances power transformer fault diagnosis by integrating multi-modal data, including dissolved gas analysis (DGA) and infrared images, to improve accuracy and efficiency. MULAN~\cite{zheng2024mulan} converts log sequences into time-series data using a log-tailored language model and employs contrastive learning to leverage multi-modal data, facilitating root-cause discovery for system failures. 
In computer vision, LA-GCN~\cite{lagcn} utilizes textual embeddings of joint names and action labels to generate faithful structural priors, enhancing skeleton-based action modeling and improving recognition tasks. In speech applications, AV-HuBERT \cite{shi2022learning} employs a self-supervised representation learning framework to leverage correlated audio and visual information \cite{zhang2023speechgpt}, while SpeechGPT~\cite{shi2022learning,zhang2023speechgpt} integrates audio and text to enhance generation performance.

\subsection*{Spatial Time Series}
\subsection{Transportation and Mobility}
Several recent studies on traffic prediction highlight the importance of multi-modal contexts to enhance forecasting accuracy. Guo  \textit{et al.} ~\cite{guo2024towards} transforms California traffic data into structured LLM prompts. 
The method uses LLaMA models and instruction fine-tuning to improve spatial-temporal learning. Meanwhile, Li \textit{et al.} ~\cite{Li2024UrbanGPTSL} employs a spatial-temporal dependency encoder to align numerical New York City traffic data with LLMs, incorporating weather, geographic context, and historical flow patterns to refine predictions. Similarly, Feng \textit{et al.} ~\cite{feng2024citygpt} proposes CityGPT, enhancing LLMs' spatial cognition for urban reasoning, mobility prediction, and navigation by integrating urban mobility data, road networks, and human behavior through instruction tuning. These studies demonstrate that LLM-based multi-modal fusion not only enhances traffic forecasting but also improves model interpretability and adaptability across diverse urban scenarios.

\subsection{Environment}
Integrating multi-modal information benefits environmental studies, particularly by addressing the prevalent challenge of missing values. VIMTS~\cite{zhao2022vimts} utilizes a structured variational approximation technique to impute missing high-dimensional modalities (stream image data) by transforming them into low-dimensional features derived from simpler, related modalities (meteorological time series records), ensuring cross-modal correlations and interpretability of the imputation process. Additionally, LITE~\cite{li2024lite} addresses the imputation of missing features through a sparse Mixture of Experts framework. It integrates and encodes various environmental variables through a unified encoder. Directed by domain-specific instructions, a language model is utilized to merge these multi-modal representations, thereby improving the accuracy of environmental spatial-temporal predictions.

\section{Future Research Directions}
In this section, we outline several underexplored research directions that open up opportunities for future advancements.

\vspace{0.2cm}

\noindent \textbf{Reasoning with Multi-modal Time Series.} 
Enhancing reasoning with multi-modal time series is pivotal for the development of intelligent systems. Future research should focus on creating a unified framework that can seamlessly integrate temporal reasoning with contextual understanding, enabling models to handle multiple time series tasks with interpretability. One potential path is to incorporate external knowledge bases and real-world context, such as developing a retrieval-augmented generation (RAG)~\cite{ning2025TSRAG} system, to enhance the reasoning process and allow models to make informed inferences beyond the immediate data. It is also promising to synergize time series model and language agents to provide more faithful and reliable reasoning on real-world contexts~\cite{shen2024exploringmultimodalintegrationtoolaugmented,jiang2025explainablemultimodaltimeseries}. The recent development of LLM reasoning models, such as chain of thoughts~\cite{chainofthought} and tree of thoughts~\cite{yao2023tree}, also offers potential solutions to improve reasoning quality.

\vspace{0.1cm}

\noindent \textbf{Decision Making.} 
Multi-modal time series analysis presents a promising future direction to enhance decision-making processes, which is crucial in high-stakes applications. By leveraging predictive signals and explanations from multi-modal contexts, future research can develop more adaptive, interpretable, and reliable decision-support systems, to facilitate the downstream optimization tasks such as resource allocation and risk management.

\vspace{0.1cm}

\noindent \textbf{Domain Generalization.}
One key challenge in multi-modal time series analysis is domain generalization, which enables a model trained on one or more source domains to effectively generalize to unseen target domains, ensuring robustness against distribution shifts. In multi-modal time series, distribution shifts can be multifaceted, stemming not only from 
time series, but also from other modalities. Therefore, it is crucial to develop specialized domain generalization methods for effective multi-modal time series analysis, including strategies to identify and preserve domain-invariant components across modalities while capturing modality-specific variations for rapid adaptation. Additionally, disentangling the effects of each modality is essential to better understand their individual contributions and mitigate cross-modal interference.

\vspace{0.1cm}

\noindent \textbf{Robustness to Missing and Noisy Modalities.} Multi-modal time series analysis often  frequently encounters messy real-world contexts with incomplete or noisy data.  Existing methods employ an iterative context-refinement algorithm \cite{jiang2025explainablemultimodaltimeseries} that filters out less relevant information, thereby enhancing the predictive insights derived from multi-modal time series.
Nonetheless, effectively dealing with missing and noisy modalities still demands further exploration. In particular, developing strategies for modality-specific imputation, noise reduction, and relevance quantification will be crucial to improving the real-world applicability of existing multi-modal time series methods.

\vspace{0.1cm}

\noindent \textbf{Ethical Considerations and Bias Mitigation.} In light of potential biases in multi-modal time series datasets, future research should integrate fairness-aware techniques, such as fairness constraints, counterfactual analysis, and adversarial debiasing. These methods should be combined with robust bias assessment frameworks to systematically detect and mitigate inequities, ensuring outcomes that are both equitable and socially responsible.

\section{Conclusion}
In this survey, we provide a comprehensive overview of existing multi-modal time series methods. We first discuss the multi-modal time series used in existing methods. Then, we propose a taxonomy based on cross-modal interactions between time series and other modalities. The existing methods are categorized and summarized accordingly. We also discuss the real-world applications and highlight future research directions in this promising area.


\begin{acks}
This research was supported in part by the National Science Foundation (NSF) CAREER IIS-2338878, as well as by generous research gifts from NEC Labs America Inc. and Morgan Stanley.
\end{acks}

\bibliographystyle{ACM-Reference-Format}
\bibliography{main}



\end{document}